\documentclass{article}

\usepackage{microtype}
\usepackage{graphicx}
\usepackage{subfigure}
\usepackage{booktabs} 

\usepackage{hyperref}

\usepackage[accepted]{icml2024}

\usepackage{amsmath}
\usepackage{amssymb}
\usepackage{mathtools}
\usepackage{amsthm}

\usepackage{enumitem}
\usepackage{xcolor}
\usepackage{graphicx} 
\usepackage{multirow} 

\usepackage{multirow}
\usepackage{graphicx}
\usepackage{rotating}
\usepackage{array}

\usepackage[capitalize,noabbrev]{cleveref}

\theoremstyle{plain}

\theoremstyle{definition}

\theoremstyle{remark}

\usepackage[textsize=tiny]{todonotes}

\icmltitlerunning{Self-Supervised Learning for Identifying Defects in Sewer Footage}

\begin{document}

\twocolumn[
\icmltitle{Self-Supervised Learning for Identifying Maintenance Defects in Sewer Footage}

\icmlsetsymbol{equal}{*}

\begin{icmlauthorlist}
\icmlauthor{Daniel Otero - EXIT83}{equal,comp}
\icmlauthor{Rafael Mateus - EXIT83}{equal,comp}
\end{icmlauthorlist}

\icmlaffiliation{comp}{EXIT83 LLC Consulting Services, Seattle, United States}

\icmlcorrespondingauthor{Daniel Otero}{daniel@exit83.com}
\icmlcorrespondingauthor{Rafael Mateus}{rafael@exit83.com}

\icmlkeywords{Machine Learning, ICML}

\vskip 0.3in
]

\printAffiliationsAndNotice{\icmlEqualContribution}

\begin{abstract}

Sewerage infrastructure is among the most expensive modern investments requiring time-intensive manual inspections by qualified personnel. Our study addresses the need for automated solutions without relying on large amounts of labeled data. We propose a novel application of Self-Supervised Learning (SSL) for sewer inspection that offers a scalable and cost-effective solution for defect detection. We achieve competitive results with a model that is at least 5 times smaller than other approaches found in the literature and obtain competitive performance with 10\% of the available data when training with a larger architecture. Our findings highlight the potential of SSL to revolutionize sewer maintenance in resource-limited settings.

\end{abstract}

\section{Introduction}
\label{submission}

The high expenses and labor-intensive process of gathering labeled data have driven researchers to seek innovative methods to train neural networks without annotations or with minimal annotated data. Self-Supervised Learning (SSL) emerges as an unsupervised learning strategy in which models learn to understand and represent data using their structure as the supervision signal \citep{ozbulak2023know}. The application of SSL techniques to computer vision has revolutionized the field, not only pushing the boundaries of unsupervised pretraining performance on popular benchmarks, such as ImageNet \citep{deng2009imagenet}, but also leading researchers to adapt these methods to effectively tackle domain-specific challenges. 
   
Sewerage infrastructure is one of the most costly in modern society, with traditional manual inspections required to identify defects. This process is limited by the number of qualified personnel and the time it takes to inspect each pipe \citep{harum2021sewer}. Given these limitations, adopting an automated approach is both practical and necessary. However, the success of these methods depends on the availability of large amounts of labeled data, which is difficult to collect due to the shortage of inspectors. We recognize the necessity to create automated solutions without the need for vast amounts of labeled data.

We are the first to propose applying self-supervised learning to the domain of sewer infrastructure inspection. We introduce a straightforward approach that uses the DINO methodology that achieves competitive results with state-of-the-art methods without the need for complex implementations. Our approach not only demonstrates the adaptability of SSL in a specialized field but also sets the groundwork for future innovations in maintaining critical urban infrastructure.

We evaluate our approach on the Sewer-ML dataset \citep{harum2021sewer}, a multi-label dataset that contains 1.3 million images and 17 different types of defects. This study demonstrates strong results (50.05 F2$_{CIW}$ and 87.45 F1$_{Normal}$) when fine-tuning with only 10\% of the available data, significantly reducing the need for annotations. Additionally, we successfully trained a much smaller model compared to state-of-the-art methods, making it ideal for deployment on small devices for live detection and enhancing scalability in resource-limited settings.

\section{Related work}
\label{sec:problem}

\textbf{Self-supervised learning.} \quad SSL methods can be broadly categorized as contrastive or non-contrastive based on how they avoid representation collapse \citep{balestriero2023cookbook, ozbulak2023know}. Contrastive methods use positive and negative pairs to help the model distinguish between different instances by comparing similar and dissimilar examples \citep{chen2020asimple, he2020momentum}. On the other hand, non-contrastive methods avoid explicit negative pairs and use strategies like clustering \citep{caron2020unsupervised}, distillation \citep{caron2021emerging}, redundancy reduction \citep{bardes2022vicreg}, or masked image modeling \citep{assran2022masked, assran2023self} to ensure rich feature extraction.

Among the non-contrastive distillation methods, we highlight DINO \citep{caron2021emerging} as it is part of our methodology. Self-distillation involves a teacher network generating pseudo-labels that a student network aims to replicate, encouraging the student to learn robust representations. The student and teacher networks share the same architecture and the teacher parameters are updated using an exponential moving average of the student ones, providing stable targets and preventing the model from collapsing to trivial solutions.
We explain in detail how DINO is used within our approach in Section \ref{sec:methodology}.

Recent research on the application of self-supervision to domain-specific tasks has shown encouraging results.  For instance, SSL has achieved state-of-the-art performance in pixel-wise anomaly localization \citep{li2021cutpaste}. Moreover, SSL has matched and surpassed the performance of clinical experts in medical imaging \citep{zhang2023dive, Azizi2023MedImaging}, has demonstrated superior performance in 3D facial image texture reconstruction \citep{ZENG2021-3Dface}, and has successfully addressed label deficiencies in training the backbone network for an RGB-D object tracking problem \citep{ZHU2024-RGBDtracking}. 

\textbf{Sewer-ML literature.} \quad The Sewer-ML benchmark introduced state-of-the-art graph-based models such as KSSNet \citep{wang2020multi}, as well as popular vanilla architectures like ResNet-101 \citep{wu2019tencent} and TResNet \citep{ridnik2020tresnet} (see Table \ref{tab:sota-metrics}). Despite their different methodologies, these approaches achieve very similar performance.

Seeking to improve the presented baseline, \citet{hauru2022multi} proposed using a hybrid vision transformer combined with a Sinkhorn tokenizer (HViT-Sk). This method enhances model efficiency and accuracy by using CNN-generated feature maps as inputs to the ViT \cite{dosovitskiy2020animage} and employing the Sinkhorn tokenizer to eliminate redundancies. Building on this, they later proposed a multi-task learning approach (CT-GAT), where a common backbone network is jointly optimized by multiple task-specific GNN heads, resulting in a more robust and versatile inspection system \citep{haurum2022multi}.

Moreover, \citet{tao2022cafen} combine features extracted by a graph-based module and a CNN with block attention modules. The graph-based module is used to capture the correlation information between labels. Similarly, \citet{hu2023toward} worked on maximizing the defect-relevant information. They proposed a Self-Purification Module (SPM) that splits the feature representation space into the sum of two spaces: defect-relevant and defect-irrelevant features. They optimized the network using three loss terms: one to purify defect-relevant features, one to decorrelate defect-irrelevant features, and one to prevent collapse. Furthermore, \citet{zhao2022towards} used Bayesian techniques to train an ``uncertainty-aware'' neural network (TMSDC). The main objective is that the model learns to ``know the unknown'' so it avoids making over-confident predictions on under-represented observations.

\begin{table}[t]
\caption{\textbf{Comparison with methods found on Sewer-ML literature.} We present experiments with ViT-T/16 and ViT-S/16 using 100\% of the data for fine-tuning. Our approaches use smaller and thus more compute-efficient architectures.}
\label{tab:sota-metrics}
\vskip 0.15in
\begin{center}
\begin{small}
\begin{sc}
\resizebox{\columnwidth}{!}{
\begin{tabular}{clccc}
\toprule
& Method & Parameters & F2$_{CIW}$ (\%) & F1$_{Normal}$ (\%) \\
\midrule
\parbox[t]{2mm}{\multirow{8}{*}{\rotatebox[origin=c]{90}{Literature}}}
& ResNet101               & 42.5M  & 53.26 & 79.55 \\
& KSSNet               & 45.2M  & 54.42 & 80.60 \\
& TResNet-L             & 53.6M & 54.63 & 81.22 \\
& TResNet-L+TMSDC       & 53.6M  & 54.54 & 81.15 \\
& CT-GAT                  & 24M    & 61.70 & 91.94 \\
& ResNet-50-HViT-Sk & 25.3M  & 60.42 & \textbf{92.41} \\
& TResNet-L+SPM         & 53.6M  & \textbf{63.38} & 91.57 \\
\midrule
\parbox[t]{2mm}{\multirow{2}{*}{\rotatebox[origin=c]{90}{Ours}}} & ViT-T/16-100\%         & \textbf{5.5M}   & 58.18 & 89.76 \\
& ViT-S/16-100\%     & \textbf{21.6M} & 60.39 & 90.13 \\
\bottomrule
\end{tabular}
}
\end{sc}
\end{small}
\end{center}
\vskip -0.1in
\end{table}

Although our results do not surpass the state-of-the-art, they provide competitive performance with much smaller architectures, providing a low-compute, cost-efficient methodology, reducing data-labeling costs and improving scalability.

\section{Methodology}
\label{sec:methodology}

\subsection{Standard approach to SSL}

In computer vision, self-supervised learning teaches neural networks to understand images using unlabeled data. This is accomplished by generating multiple random augmentations of the same image and training the model to recognize that these different views all originate from the same source. This is referred to as the pretext task and aims to teach the model to generate similar embeddings for similar inputs and dissimilar embeddings for dissimilar ones.

\textbf{Mathematical definition.} \quad Let $f_{\theta}$ be an encoder \textit{backbone} with parameters $\theta$ that produces vector representations $r$ from augmented views $x_t$ of an image $x$ produced by a stochastic function $\mathbb{T}(x) = x_t$. Representations $r$ can be mapped to projections $z$ and predictions $z'$ using projector $g_{\gamma}$ and predictor $q_{\tau}$ functions, where $g_{\gamma}(f_{\theta}(x)) = z$ and $q_{\tau}(g_{\gamma}(f_{\theta}(x))) = z'$. In this context $g_{\gamma}$ and $q_{\tau}$ are MLPs. 

Like other popular self-supervised approaches \citep{chen2020asimple, grill2020bootstrap, caron2020unsupervised, bardes2022vicreg}, DINO employs a projection head on top of the encoder backbone, with the loss being computed on the projector's output. The projector function acts as an informational bottleneck, ensuring that the backbone's representations are not overly biased to merely comply with the self-supervised learning objective \citep{chen2020asimple}. 

This comprises the intuition behind self-supervised pretraining. For evaluating performance on downstream tasks, only the encoder backbone from the pretraining phase is retained. Afterwards, a labeled dataset is used to either fine-tune the model or train a linear classifier on top of the frozen backbone.

\subsection{Implementation details}

\textbf{Architecture.} \quad For the self-supervised pretraining, we used the DINO methodology. For the encoder backbones, we used the ViT Tiny (ViT-T/16) and ViT Small (ViT-S/16) models, which primarily differ in the number of parameters—5.5M and 21.6M respectively—and computational complexity, with ViT-T/16 having 192 hidden layers and 3 heads, and ViT-S/16 having 384 hidden layers and 6 heads.

The projector of the models comprised an MLP with two hidden layers of size 2048 and an output layer of size 256. The loss was computed with respect to 32,768 prototypes. For other DINO hyperparameters, we adhered to the recommendations in the original paper \citep{caron2021emerging}. The training was performed using Pytorch 2.0.2 \citep{paszke2019pytorch} on 16 Tesla T4 GPUs, using the maximum batch size that could fit into memory for each model. Our code development was greatly inspired by the solo-learn library \citep{turrisi2022solo-learn}.

\textbf{Global views instead of multi-crop.} \quad Sewer-ML is a multi-label dataset where defects vary in shape and size. To avoid matching local views with fewer defects (or none) to global views containing the full image, we did not perform multi-crop. This decision was made to prevent potential mismatches in embeddings and to avoid hindering the neural network optimization during pretraining.

\textbf{Optimization.} \quad The experiments for pretraining were conducted over 35 epochs using the AdamW optimizer. The base learning rate was set to \(5 \times 10^{-5} \times \text{batch\_size} / 256\). A linear warmup starting at \(3 \times 10^{-5}\) was applied for the first 10 epochs, followed by a cosine scheduler with no restarts. The base and final decay rates (\(\tau\)) were 0.996 and 0.999, respectively, with a minimum learning rate of \(1 \times 10^{-6}\).

For fine-tuning, we took the pretrained backbone and placed an untrained classifier head on top of it. The experiments were run for 45 epochs using the AdamW optimizer, with a base learning rate of \(5 \times 10^{-4} \times \text{batch\_size} / 256\). A multistep scheduler with a gamma of 0.1 was used, with step milestones at epochs 15 and 35.

\textbf{Loss function and positive weights.} \quad Given the unbalanced nature of the dataset and the superior importance of recall over precision in the benchmark metrics, it is necessary to craft a custom-weighted loss to effectively address the task. We optimized the model with respect to a binary cross-entropy loss with positive weighting. The coefficients were built based on the class importance values proposed in the benchmark and were calculated using the following formula:

$$
pos\_weight_c = 2 \times \left( 1 + \frac{CIW_c}{\frac{1}{C} \sum_{c=1}^{C} CIW_c} \right)
$$

The motivation behind this formula is to first normalize each class's importance value by dividing it by their mean. This provides insight into how significant each class is relative to the overall distribution. Subsequently, we add 1 to this term to place greater emphasis on the positive samples, then multiply by 2 to further enhance the emphasis.

\subsection{Sewer-ML benchmark metrics}

To assess the performance of the multi-label benchmark, we use the proposed metrics. A weighted F2 metric (F2$_{CIW}$) for defect prediction and a regular F1 score (F1$_{Normal}$) for non-defect predictions \citep{harum2021sewer}. The weights for the F2 metric are assigned to each defect class based on their economic impact. Moreover, the F2 score is employed to prioritize recall over precision since missing a defect has a greater economic impact than generating a false positive.

\section{Results}
\label{sec:results}

We conducted several experiments to evaluate our models. These experiments include reporting metrics for the pretrained architectures by (i) training a linear classifier on top of the frozen backbone, (ii) fine-tuning the models using 10\%, 50\%, and 100\% of the data, and (iii) pretraining the models using a hybrid approach that incorporates both self-supervised and supervised losses. For comparison purposes, we also trained the models in a fully supervised setting. All experiments were performed using the ViT-T/16 and ViT-S/16 architectures.

\begin{table}[t]
\caption{\textbf{Performance comparison with varying data sizes.} This table presents a comparison in performance between the proposed SSL approach and a fully supervised setting across different data sizes (10\%, 50\%, and 100\% of the total dataset) for the ViT-T/16 and ViT-S/16 models. }

\label{tab:experimentation}
\vskip 0.15in
\begin{center}
\begin{small}
\begin{sc}
\resizebox{\columnwidth}{!}{
\begin{tabular}{lcccccc}
\toprule
& \multicolumn{2}{c}{\textbf{SSL + Finetuning}} & \multicolumn{2}{c}{\textbf{Fully Supervised}} \\
\cmidrule(lr){2-3} \cmidrule(lr){4-5}
Method & F2$_{CIW}$ (\%) & F1$_{Normal}$ (\%) & F2$_{CIW}$ (\%) & F1$_{Normal}$ (\%) \\
\midrule
ViT-T-16-hybrid    &  37.95 & 80.96 & - & -\\
ViT-T/16-linear   & 25.84 & 57.04 & - & - \\
ViT-T/16-10\% & 28.58 & 82.14 & 32.65 & 82.29 \\
ViT-T/16-50\% & 52.78 & 88.32 & 50.15 & 87.60 \\
ViT-T/16-100\% & 58.18 & \textbf{89.76} & \textbf{58.94} & 89.68 \\
\midrule
ViT-S/16-hybrid    & 43.48 & 86.54 & - & - \\
ViT-S/16-linear    & 30.87 & 62.65 & - & - \\
ViT-S/16-10\% & 50.05 & 87.45 & 36.44 & 83.48 \\
ViT-S/16-50\% & 57.17 & \textbf{90.18} & 56.23 & 88.60 \\
ViT-S/16-100\% & \textbf{60.39} & 90.13 & 58.81 & 89.95 \\
\bottomrule
\end{tabular}
}
\end{sc}
\end{small}
\end{center}
\vskip -0.1in
\end{table}

\textbf{Performance.} \quad Our experiments with the ViT-S model demonstrate its robustness across varying data levels. When using 100\% of the data for fine-tuning, its performance was on par with state-of-the-art methods. Using 50\% of the data, ViT-S performed nearly as well as when using the full dataset. Even with just 10\% of the data, the model showed solid baseline performance, proving effective in data-scarce scenarios (see Table \ref{tab:experimentation}). For both architectures, the hybrid approach enhanced non-defect detection but demonstrated limited performance for identifying defects. We hypothesize that the self-supervised signal enabled the model to encode richer representations of non-defective pipes. However, this also limited the feature exploitation of the supervised loss, affecting defect detection results.

The findings underscore the competitive performance of our proposed self-supervised learning approach with fine-tuning compared to fully supervised learning. While the fully supervised method achieves slightly higher metrics in smaller architectures (ViT-T/16) with 10\% of the data for fine-tuning, the SSL method shows substantial improvements with increased model complexity, surpassing the performance of all ViT-S/16 configurations.

\textbf{Parameter count efficiency.} \quad Our approach significantly reduces the size of the networks required for training while maintaining effective performance. While some state-of-the-art methods exceed 50 million parameters, our largest model has approximately 21.6 million, achieving similar results with around half the size. Moreover, using the ViT-T model, we obtained satisfactory outcomes even when fine-tuning on just 50\% of the data, achieving similar performance to the approaches proposed in the original paper but with a model at least 9 times smaller. Furthermore, fine-tuning the ViT-T on the whole dataset yields very similar results to the ones obtained by fine-tuning ViT-S on 50\% of the data, demonstrating the effectiveness of our approach even with smaller models.

\textbf{Simplicity and effectiveness of the approach.} \quad Current methods often require specialized knowledge and extensive labeled data. In contrast, our approach is straightforward, involving only pretraining and fine-tuning, which are standard practices in transfer learning, as well as requiring significantly fewer labels due to our use of self-supervision methodologies. This simplicity not only makes our method more accessible but also offers greater adaptability, allowing for effective performance with less labeled data while still achieving comparable results to more complex methods.

\textbf{Informational content.} \quad We employed the RankMe metric \citep{garrido2023rankme} to monitor the informational content of representations during pretraining. A higher value suggests greater informational content. Results showed that hybrid signals had significantly lower semantic content (see Table \ref{tab:rankme}), validating that self-supervision produces richer representations, whereas supervised methods primarily exploit local features. Furthermore, the ViT-S demonstrated a lower informational content than ViT-T when pretrained in a self-supervised manner. We presume that this is due to the absence of multi-crop, which acts as a regularizer for larger models \citep{tan2023effective}. 

\begin{table}[t]
\caption{\textbf{RankMe values.} Final values gathered during training.}
\label{tab:rankme}
\vskip 0.15in
\begin{center}
\begin{small}
\begin{sc}
\begin{tabular}{lcccr}
\toprule
Method & RankMe \\
\midrule
ViT-T/16            & 74.37 \\
ViT-T/16 Hybrid     & 26.71 \\
ViT-S/16           & 50.56 \\
ViT-S/16 Hybrid    & 26.87 \\
\bottomrule
\end{tabular}
\end{sc}
\end{small}
\end{center}
\vskip -0.1in
\end{table}

\section{Conclusions}
\label{sec:conclusions}

Our research demonstrates the effective application of self-supervised learning to the domain of sewer infrastructure inspection, specifically in defect detection, a field traditionally reliant on labor-intensive and costly manual inspections. This approach not only achieves high-performance results with minimal labeled data but also provides a scalable and cost-effective solution for urban infrastructure maintenance.

Even when fine-tuning with only 10\% of the available data, our research achieves notable results. We propose deploying a smaller model in production—approximately 20\% the size of state-of-the-art models—that delivers robust performance. This approach reduces the need for extensive labeling and optimizes model size for on-device scalability in live detection. Although not the primary focus of this study, we observed that the ViT-T/16 model performs well in a fully supervised setting, which is a promising result considering its compact architecture. 

For future research, it is essential to investigate the potential of various self-supervised learning methods that have not yet been applied to sewer infrastructure inspection, particularly by assessing their performance in low-data, low-compute environments. While Sewer-ML is a curated dataset, it may not fully reflect the complexities of real sewer inspections, particularly the defect-to-non-defect ratio. Therefore, the proposed method might not be immediately applicable out-of-the-box and may require extensive experimentation with other self-supervised learning techniques. Nevertheless, training a foundational model on sewer pipes offers the novel potential for transferability to a broader range of tasks within this industry.

\section*{Acknowledgements}
This project was funded by EXIT83 Consulting LLC. We extend our gratitude specially to the EXIT83 AI Team for their invaluable support and feedback, which contributed to the final version of this paper. Additionally, we appreciate the constructive comments from the anonymous reviewers of the LatinX in AI workshop.

\bibliography{main}
\bibliographystyle{icml2024}

\newpage
\appendix
\onecolumn
\section{Image Augmentations}
\quad During self-supervised pretraining, we employed several augmentations to enhance the diversity of the training dataset. Specifically, we applied random crops and resized the images to 224x224, using a scale ranging from 0.5 to 1.0 and bicubic interpolation. We applied color jitter to adjust the brightness, contrast, saturation, and hue of the images. Additionally, we included random grayscaling with a probability of 0.15, also random Gaussian blurring with a probability of 0.3 and a sigma ranging from 0.1 to 1, and finally random equalization and solarization with a probability of 0.3. Horizontal flipping was performed randomly. Finally, all images were normalized. During validation, the images were only resized and normalized.

We used a slightly different image augmentation pipeline for fine-tuning. Instead of performing random crops, we used full image resizes. We keep augmentations like color jitter, random horizontal flip, and normalization, consistent with the pretrain augmentations. We replaced the remaining transformations with random equalizing and random autocontrasting. We also incorporated random affine augmentations with a rotation limit of 5 degrees and applied random erasing with a scale ranging from 0.01 to 0.05 and a ratio ranging from 0.1 to 1. Validation augmentations remained the same as for pretraining.

\end{document}